\documentclass[12pt,twocolumn,letterpaper]{article}

\usepackage[pagenumbers]{cvpr} 

\usepackage{graphicx}
\usepackage{amsmath}
\usepackage{amssymb}
\usepackage{booktabs}

\usepackage[pagebackref,breaklinks,colorlinks]{hyperref}

\usepackage[capitalize]{cleveref}
\crefname{section}{Sec.}{Secs.}
\Crefname{section}{Section}{Sections}
\Crefname{table}{Table}{Tables}
\crefname{table}{Tab.}{Tabs.}

\begin{document}
\title{Use Classifier as Generator}
\author{Haoyang Li\\
\small Cornell University
\\
{\tt\small hl2425@cornell.edu}}
\maketitle

\begin{abstract}
  Image recognition/classification is a widely studied problem, but its reverse problem, image generation, has drawn much less attention until recently. But the vast majority of current methods for image generation require training/retraining a classifier and/or a generator with certain constraints, which can be hard to achieve. In this paper, we propose a simple approach to directly use a normally trained classifier to generate images. We evaluate our method on MNIST and show that it produces recognizable results for human eyes with limited quality with experiments. \footnote{this is a course project paper for CS 6670 Fall 2021, check  \href{https://www.youtube.com/watch?v=Kr_GAzfuscQ}{here} for a recorded presentation}
\end{abstract}

\section{Introduction \& Related Works}

Image recognition/classification is a widely studied problem in the area of computer vision. As the very focus of research, computer vision algorithms have evolved to perform on par with human over image classification benchmarks such as MNIST\cite{lecun1998gradient}, CIFAR-10\cite{krizhevsky2009learning} and ImageNet\cite{5206848}. On the other hand, image generation has drawn less attention until the proposal of variational autoencoder\cite{kingma2014autoencoding}, generative adversarial nets\cite{goodfellow2014generative} and normalizing flows\cite{rezende2015variational}. These three approaches take the current bulk of image generation research.

Relating image generation to image classification is not something new. Exploiting image generation to image classification dates way back before the era of deep learning\cite{jaakkola1999exploiting}, but the prevailing deep learning models all takes a discriminative approach for image classification\cite{krizhevsky2012imagenet}\cite{simonyan2014very}\cite{he2016deep}. Recently, generative approach gets attention again as an alternative in defense against adversarial attacks\cite{william2017safer}. 

As it is so natural to make use of image generation for image classification, it makes us curious if we could exploit image classification for image generation. And without surprise, using image classifier for image generation is not a completely unexplored idea. There are two existing approaches in the literature. 

The first approach is to make the classifier itself invertible and then inverse the classifier and turn it into a generator. Normalizing flows\cite{rezende2015variational} adopt an inherently invertible structure for classifier and then use its inverse as generator. Invertible residual networks\cite{behrmann2019invertible} enforces a strong constraint on the Lipshitz constant of each layer in a residual network to make it invertible.

The second approach is to use the classifier as a criteria for generator. Generative adversarial nets\cite{goodfellow2014generative} design a game between a generator and a discriminator, i.e. a binary classifier, use the latter as the criteria of the former and train them together. Improving generator or dicriminator yields a family of generative adversarial nets that have gained great success on image generation\cite{radford2015unsupervised}\cite{mirza2014conditional}\cite{zhu2017unpaired}.

\section{Motivation}\label{sec:motivation}

Generative adversarial nets explicitly train a generator to produce images. However, it seems that the classifier itself can supervise the generation of image alone under certain conditions as it has been reported that using a single adversarially trained image classifier as the criteria and optimize input, we can generate images of some quality\cite{santurkar2019image}. Prior to that, similar approaches are used to generate interpretations for general classifiers\cite{selvaraju2017grad} and transfer textures between images\cite{jing2019neural}.

But all of these methods require training image classifiers/generators under specific constraints, which can be hard to achieve. We cannot just take a trained classifier and use it to generate images with these methods. This brings the problem we are interested to investigate in this paper: 

\emph{Can we generate images using a normally trained classifier itself?}

\begin{figure*}
    \centering
    \includegraphics[width=140mm]{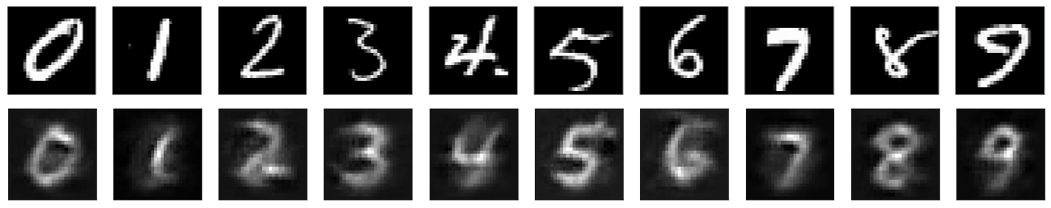}
    \caption{Reference samples (first row) and generated images (second row)}
    \label{fig:digits_demo1}
\end{figure*}


\begin{figure*}
    \centering
    \includegraphics[width=140mm]{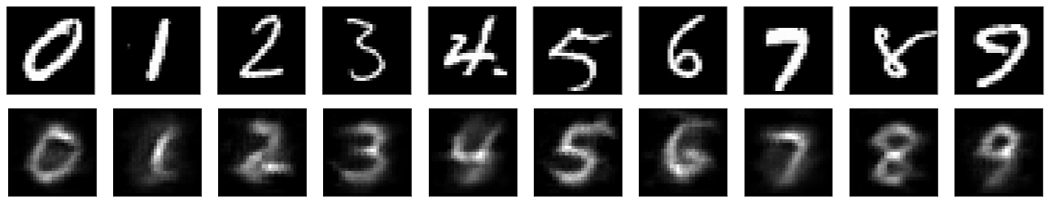}
    \caption{Reference samples (first row) and generated images using projected gradient descent (second row)}    
    \label{fig:digits_demo3}
\end{figure*}

\begin{figure*}
    \centering
    \includegraphics[width=140mm]{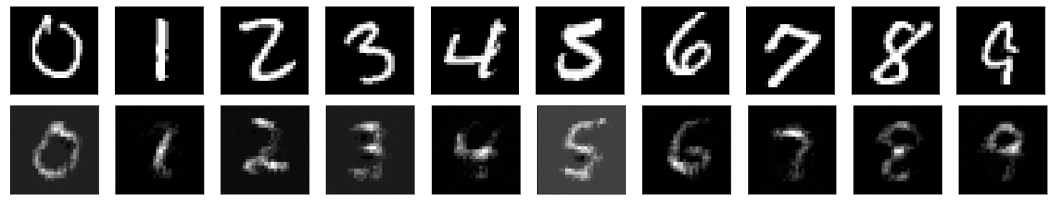}
    \caption{Reference samples (first row) and generated images using $\ell_1$-distance for regularizer (second row)}    
    \label{fig:digits_demo4}
\end{figure*}

\section{Problem Definition}

A classifier as an oracle that feeds on input and gives label cannot provide any information to guide the generation of images. To make this problem feasible, we need to make some assumptions. We first assume that the classifier produces an estimation of the probability that the fed input belongs to a certain class rather than just a label. 

Besides, we also assume it to be of reasonable performance that it can provide domain knowledge about the dataset and of course, we know which training dataset it used or at least which kind of data it was trained for. Since we wish to use the classifier itself as a generator, we require this classifier to be differentiable or at least can be well approximated by a differentiable classifier.

Given a dataset $\mathcal{D}=\{(x,y)|x\in \mathcal{X},y\in\mathcal{Y}\}$ and an image classifier trained on it, i.e. $f(x)\in[0,1]^d$ where $d=|\mathcal{Y}|$ is the number of classes and $f(x)_y\approx p(y|x)$, i.e. the $y$-th dimension of the output of this classifier serves as an approximation of how likely the input $x$ belongs to class $y$. The final prediction result will be $\hat{y}=\arg\max_{y}f(x)_y$. What we want to acquire given the classifier and some label $y$ is
\begin{align}
    \hat{x} &= \arg\max_{x\in\mathcal{X}} p(x,y) \\
    &=\arg\max_{x\in\mathcal{X}} p(y|x)p(x)\\
    &\approx \arg\max_{x\in\mathcal{X}}f(x)_y p(x)\\
    &=\arg\max_{x\in\mathcal{X}}\log f(x)_y + \log p(x)\\
    &=\arg\min_{x\in\mathcal{X}}\mathcal{L}_1(f(x),y) + \mathcal{L}_2 (x)
\end{align}

The first loss function $\mathcal{L}_1$ should give a measure of how likely the input $x$ belongs to class $y$, the more likely that $x$ belongs to class $y$, its value should be smaller. The second loss $\mathcal{L}_2$ should give a measure of how likely the input $x$ appears, the more likely that $x$ appears, its value should be smaller. The classifier can help build the first loss function, but we still need to deal with the second loss function.

\section{Method}

It is impossible to get the possibility of each input for the entire space, i.e. we do not know about $p(x)$. To resolve this issue, we further assume that $x\sim\mathcal{N}(\mu,\sigma^2)$, where we estimate the average and variance from the original dataset. Then the maximizer of $p(x)$ is equal to the minimizer of $||x-\mu||$. We can estimate $\mu$ using the dataset, i.e. $\mu\approx\frac{\sum_{x\in\mathcal{D}} x}{N}$, then the second loss function is just the distance between the average and the current input, i.e. $\mathcal{L}_2(x)=||x-\frac{\sum_{x\in\mathcal{D}} x}{N}||$.

For the first loss function, we can adopt a surrogate loss  function $\mathcal{L}$ just like what we would use when training a classifier. Incorparating a regularizing factor $\lambda\in\mathbf{R}$, the final objective will be

\begin{align}
\hat{x}=\arg\min_{x\in\mathcal{X}}\mathcal{L}(f(x),y) +\lambda ||x-\frac{\sum_{x\in\mathcal{D}} x}{N}||
\end{align}

A simple instance of it is using cross-entropy loss for the surrogate loss and $\ell_2$-distance for the regularizer.

\section{Experiments}

In this section, we will conduct an exploratory experimental analysis for the proposed method, starting from a very basic setting. 

\subsection{Image Generation}\label{sec:basic_gen}

Using MNIST\cite{lecun1998gradient} and a simple convolutional neural network with a test accuracy of 0.96, we evaluate our method using the simple instance in the previous section. The generated digit images are presented in Figure \ref{fig:digits_demo1} along with a reference sample of the same class drawn from the training dataset. Each image is generated through 100 steps of stochastic gradient descent (SGD) with a step size of 0.05 and a regularizing factor $\lambda$ of 50. The initial samples are set to $0$. We find that each generated image is of limited quality, but recognizable for human eyes as an instance for the corresponding class.



\subsection{PGD vs SGD}

PGD is short for projected gradient descent. It has the same steps with stochastic gradient descent but after each step, it projects the modified parameters back to a feasible space. PGD is used in \cite{santurkar2019image} for image generation. They project the modified input back to the valid pixel space after each step, .e.g. a hypercube $[0,1]^d$\footnote{$d$ is the dimension of data space}.

Figure \ref{fig:digits_demo3} shows the generated images using projected gradient descent with the same hyperparameters as in \ref{sec:basic_gen}. We find that using PGD makes the results look more balanced, as it clips off highly deviated values in each image. But it makes no significant difference for the overall quality.



\subsection{$\ell_1$-distance vs $\ell_2$-distance}

The basic setting in Section \ref{sec:basic_gen} uses $\ell_2$-distance for the regularizer, which roots from our assumption that these digits are drawn from a Gaussian distribution, but it may not be the real case. Figure \ref{fig:digits_demo4} shows the generated images using $\ell_1$-distance for the regularizer, with which we now assume these digits are drawn from a Laplacian distribution. We find that the generated images are less blurred, but it does not make much difference in terms of overall quality.

\subsection{Image Interpolation}

\begin{figure*}
    \centering
    \includegraphics[width=140mm]{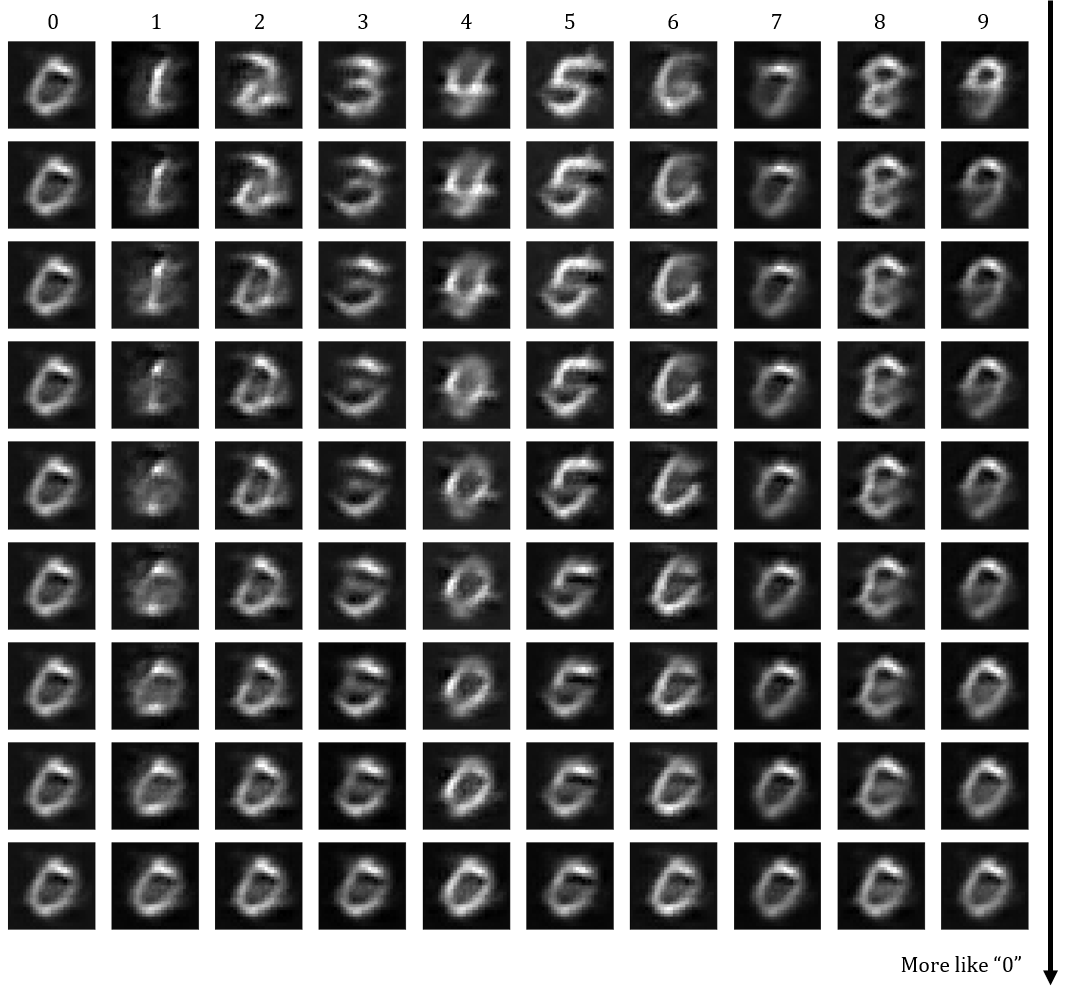}
    \caption{Image interpolations between "0" and all digits with various ratios.}
    \label{fig:image_interp}
\end{figure*}

It is also possible to interpolate images using this approach. Figure \ref{fig:image_interp} shows a series of interpolations between all digits and "0". These results are acquired with most of basic settings from Section \ref{sec:basic_gen} but a direct loss rather than a surrogate loss. For each generated image, we directly maximize its corresponding logit for its class along with the logit for class "0" and vary their ratios to get different levels of interpolations. The quality of these interpolations is not good, but we find that it does show a continuous interpolation between different digits.

\subsection{Effect of Regularization}

\begin{figure*}
    \centering
    \includegraphics[width=140mm]{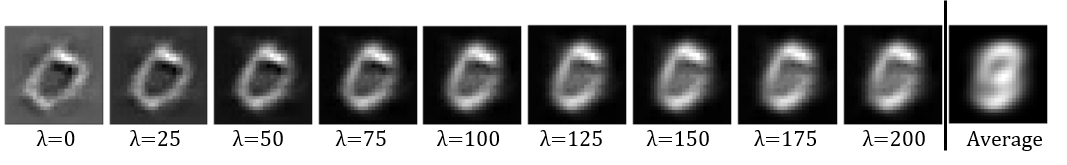}
    \caption{Effect of regularization}
    \label{fig:effect_lamb}
\end{figure*}

Figure \ref{fig:effect_lamb} shows the generated images using the basic setting in Section \ref{sec:basic_gen} with various regularizing factor $\lambda$. We expect that regularization will make the generated image look more like the average of all samples, and the results align well with our expectations. With a stronger regularization, the generated images are more plausible and look more like real samples drawn from the dataset. But when the regularization is too strong, it makes generated images look like the average.

\subsection{Effect of Classifier}

\begin{figure*}
    \centering
    \includegraphics[width=140mm]{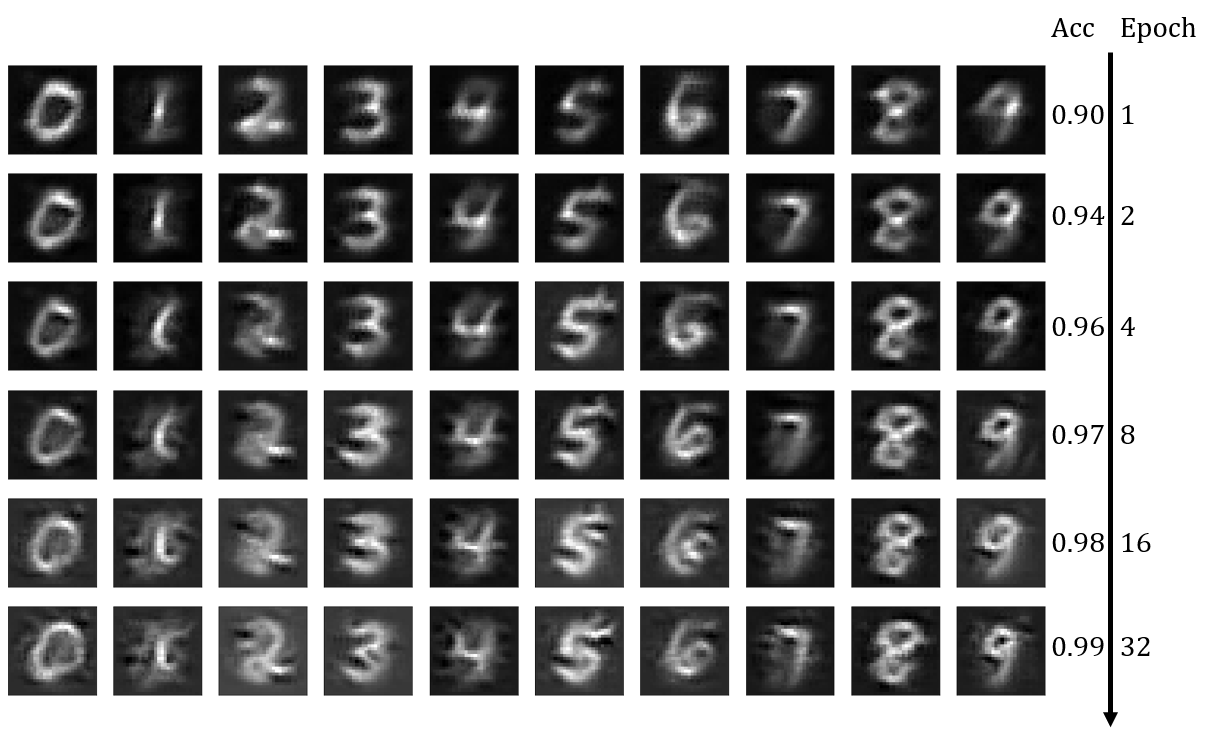}
    \caption{Generated images with classifiers trained for different epochs.}
    \label{fig:effect_epoch}
\end{figure*}

Under the basic setting used in Section \ref{sec:basic_gen}, Figure \ref{fig:effect_epoch} shows the generated images with the same classifier trained for different epochs, along with its test accuracy. We find that a classifier with higher accuracy does not always produce images of higher quality. They all produce a recognizable results, but with classifiers trained for too much epochs, the results get noisy.

\section{Conclusion}

In this paper, we propose a method to generate images using a normally trained classifier. But the method proposed here is straight, simple and not competitive with state-of-the-art methods in terms of the quality of generated images. We demonstrate with experiments that this method has some potentials for image generation, but it still needs much improvements. We also discover that the quality of the generated images is not proportional to the performance of classifier in contradictory to our intuition. It seems worth further investigations.
Due to the limited computational resources, we only evaluate it on MNIST dataset. Whether it generalizes to more complex and larger datasets or not is still questionable.


\nocite{*}
{
    \small
    \bibliographystyle{ieee_fullname}
    \bibliography{macros,main}

\begin{thebibliography}{10}\itemsep=-1pt

\bibitem{behrmann2019invertible}
Jens Behrmann, Will Grathwohl, Ricky~TQ Chen, David Duvenaud, and
  J{\"o}rn-Henrik Jacobsen.
\newblock Invertible residual networks.
\newblock In {\em International Conference on Machine Learning}, pages
  573--582. PMLR, 2019.

\bibitem{5206848}
Jia Deng, Wei Dong, Richard Socher, Li-Jia Li, Kai Li, and Li Fei-Fei.
\newblock Imagenet: A large-scale hierarchical image database.
\newblock In {\em 2009 IEEE Conference on Computer Vision and Pattern
  Recognition}, pages 248--255, 2009.

\bibitem{goodfellow2014generative}
Ian Goodfellow, Jean Pouget-Abadie, Mehdi Mirza, Bing Xu, David Warde-Farley,
  Sherjil Ozair, Aaron Courville, and Yoshua Bengio.
\newblock Generative adversarial nets.
\newblock {\em Advances in neural information processing systems}, 27, 2014.

\bibitem{he2016deep}
Kaiming He, Xiangyu Zhang, Shaoqing Ren, and Jian Sun.
\newblock Deep residual learning for image recognition.
\newblock In {\em Proceedings of the IEEE conference on computer vision and
  pattern recognition}, pages 770--778, 2016.

\bibitem{jaakkola1999exploiting}
Tommi~S Jaakkola, David Haussler, et~al.
\newblock Exploiting generative models in discriminative classifiers.
\newblock {\em Advances in neural information processing systems}, pages
  487--493, 1999.

\bibitem{jing2019neural}
Yongcheng Jing, Yezhou Yang, Zunlei Feng, Jingwen Ye, Yizhou Yu, and Mingli
  Song.
\newblock Neural style transfer: A review.
\newblock {\em IEEE transactions on visualization and computer graphics},
  26(11):3365--3385, 2019.

\bibitem{kingma2014autoencoding}
Diederik~P Kingma and Max Welling.
\newblock Auto-encoding variational bayes, 2014.

\bibitem{krizhevsky2009learning}
Alex Krizhevsky, Geoffrey Hinton, et~al.
\newblock Learning multiple layers of features from tiny images.
\newblock 2009.

\bibitem{krizhevsky2012imagenet}
Alex Krizhevsky, Ilya Sutskever, and Geoffrey~E Hinton.
\newblock Imagenet classification with deep convolutional neural networks.
\newblock {\em Advances in neural information processing systems},
  25:1097--1105, 2012.

\bibitem{lecun1998gradient}
Yann LeCun, L{\'e}on Bottou, Yoshua Bengio, and Patrick Haffner.
\newblock Gradient-based learning applied to document recognition.
\newblock {\em Proceedings of the IEEE}, 86(11):2278--2324, 1998.

\bibitem{mirza2014conditional}
Mehdi Mirza and Simon Osindero.
\newblock Conditional generative adversarial nets.
\newblock {\em arXiv preprint arXiv:1411.1784}, 2014.

\bibitem{radford2015unsupervised}
Alec Radford, Luke Metz, and Soumith Chintala.
\newblock Unsupervised representation learning with deep convolutional
  generative adversarial networks.
\newblock {\em arXiv preprint arXiv:1511.06434}, 2015.

\bibitem{rezende2015variational}
Danilo Rezende and Shakir Mohamed.
\newblock Variational inference with normalizing flows.
\newblock In {\em International conference on machine learning}, pages
  1530--1538. PMLR, 2015.

\bibitem{santurkar2019image}
Shibani Santurkar, Dimitris Tsipras, Brandon Tran, Andrew Ilyas, Logan
  Engstrom, and Aleksander Madry.
\newblock Image synthesis with a single (robust) classifier.
\newblock {\em arXiv preprint arXiv:1906.09453}, 2019.

\bibitem{selvaraju2017grad}
Ramprasaath~R Selvaraju, Michael Cogswell, Abhishek Das, Ramakrishna Vedantam,
  Devi Parikh, and Dhruv Batra.
\newblock Grad-cam: Visual explanations from deep networks via gradient-based
  localization.
\newblock In {\em Proceedings of the IEEE international conference on computer
  vision}, pages 618--626, 2017.

\bibitem{simonyan2014very}
Karen Simonyan and Andrew Zisserman.
\newblock Very deep convolutional networks for large-scale image recognition.
\newblock {\em arXiv preprint arXiv:1409.1556}, 2014.

\bibitem{william2017safer}
William Wang, Angelina Wang, Aviv Tamar, Xi Chen, and Pieter Abbeel.
\newblock Safer classification by synthesis.
\newblock {\em CoRR}, abs/1711.08534, 2017.

\bibitem{zhu2017unpaired}
Jun-Yan Zhu, Taesung Park, Phillip Isola, and Alexei~A Efros.
\newblock Unpaired image-to-image translation using cycle-consistent
  adversarial networks.
\newblock In {\em Proceedings of the IEEE international conference on computer
  vision}, pages 2223--2232, 2017.

\end{thebibliography}
}



\end{document}